\documentclass[letterpaper]{article} %
\usepackage{aaai23}  %
\usepackage{times}  %
\usepackage{helvet}  %
\usepackage{courier}  %
\usepackage[hyphens]{url}  %
\usepackage{graphicx} %
\usepackage{natbib}  %
\usepackage{caption} %
\usepackage{algorithm}

\usepackage{newfloat}
\usepackage{listings}
\DeclareCaptionStyle{ruled}{labelfont=normalfont,labelsep=colon,strut=off} %
\floatstyle{ruled}
\newfloat{listing}{tb}{lst}{}
\floatname{listing}{Listing}
\usepackage{amsmath}
\usepackage{amssymb}
\usepackage{nicefrac}       %
\usepackage{microtype}      %
\usepackage{pifont}
\usepackage{multirow}
\usepackage{verbatim}
\usepackage{color,xcolor}
\usepackage{enumitem}
\usepackage{booktabs}
\usepackage{tabulary,overpic}
\usepackage[caption=false]{subfig}
\usepackage{epstopdf}
\usepackage{algorithmicx}
\usepackage{algpseudocode}
\usepackage{enumerate}
\usepackage{bm}
\usepackage{colortbl}
\usepackage{soul}
\usepackage{mathtools}

\definecolor{citecolor}{RGB}{111,111,150} %
\usepackage[pagebackref=true,breaklinks=true,colorlinks,bookmarks=false, citecolor={citecolor}]{hyperref}

\definecolor{pos}{RGB}{0,153,0}
\definecolor{neg}{RGB}{180,0,0}
\definecolor{smpos}{RGB}{255,128,0}
\definecolor{row}{RGB}{240,240,240}

\newcommand{\fref}[1]{Fig.~\ref{#1}}
\newcommand{\tref}[1]{Table~\ref{#1}}

\newcommand{\ch}{}
\newcommand{\app}{\raise.17ex\hbox{$\scriptstyle\sim$}}
\newcommand{\round}[1]{\ensuremath{\lfloor#1\rceil}}
\newlength\savewidth\newcommand\shline{\noalign{\global\savewidth\arrayrulewidth\global\arrayrulewidth1pt}\hline\noalign{\global\arrayrulewidth\savewidth}}
\newcommand{\tablestyle}[2]{\setlength{\tabcolsep}{#1}\renewcommand{\arraystretch}{#2}\centering\footnotesize}

\title{Weakly-guided Self-supervised Pretraining for\\Temporal Activity Detection}

\author{Kumara Kahatapitiya\textsuperscript{\rm 1}\thanks{work done during an internship at Wormpex AI Research.}, Zhou Ren\textsuperscript{\rm 2}, Haoxiang Li\textsuperscript{\rm 2}, Zhenyu Wu\textsuperscript{\rm 2}, Michael S. Ryoo\textsuperscript{\rm 1}, Gang Hua\textsuperscript{\rm 2}
}

\affiliations{
    \textsuperscript{\rm 1}Stony Brook University \hspace{3mm} \textsuperscript{\rm 2}Wormpex AI Research\\

}

\usepackage{bibentry}

\begin{document}

\maketitle

\begin{abstract}

Temporal Activity Detection aims to predict activity classes per frame, in contrast to video-level predictions in Activity Classification (i.e., Activity Recognition). Due to the expensive frame-level annotations required for detection, the scale of detection datasets is limited. Thus, commonly, previous work on temporal activity detection resorts to fine-tuning a classification model pretrained on large-scale classification datasets (e.g., Kinetics-400). However, such pretrained models are not ideal for downstream detection, due to the disparity between the pretraining and the downstream fine-tuning tasks. \ch{In this work, we propose a novel \textit{weakly-guided self-supervised} pretraining method for detection. We leverage weak labels (classification) to introduce a self-supervised pretext task (detection) by generating frame-level pseudo labels, multi-action frames, and action segments. Simply put, we design a detection task similar to downstream, on large-scale classification data, without extra annotations.} We show that the models pretrained with the proposed weakly-guided self-supervised detection task outperform prior work on multiple challenging activity detection benchmarks, including Charades and MultiTHUMOS. Our extensive ablations further provide insights on when and how to use the proposed models for activity detection. Code is available at \href{https://github.com/kkahatapitiya/SSDet}{\texttt{github.com/kkahatapitiya/SSDet}}.

\end{abstract}

\section{Introduction}

Pretraining has become an indispensable component in the deep learning pipeline. Most computer vision tasks leverage large-scale labeled or unlabeled data to do pretraining in a supervised or unsupervised way, which gives performance boosts in downstream tasks, especially when training data is scarce. Such benefits of pretraining have been observed in many applications including object detection \cite{mahajan2018exploring, dai2021up}, segmentation \cite{poudel2019fast}, video understanding \cite{ghadiyaram2019large}, reinforcement learning \cite{schwarzer2021pretraining} and language modeling \cite{liu2019roberta}. This behavior can be attributed to models becoming more robust by looking at more data, which helps generalize to unseen distributions in the downstream tasks \cite{bommasani2021opportunities}.

\begin{figure}[t]
	\centering
	\includegraphics[width=0.95\linewidth]{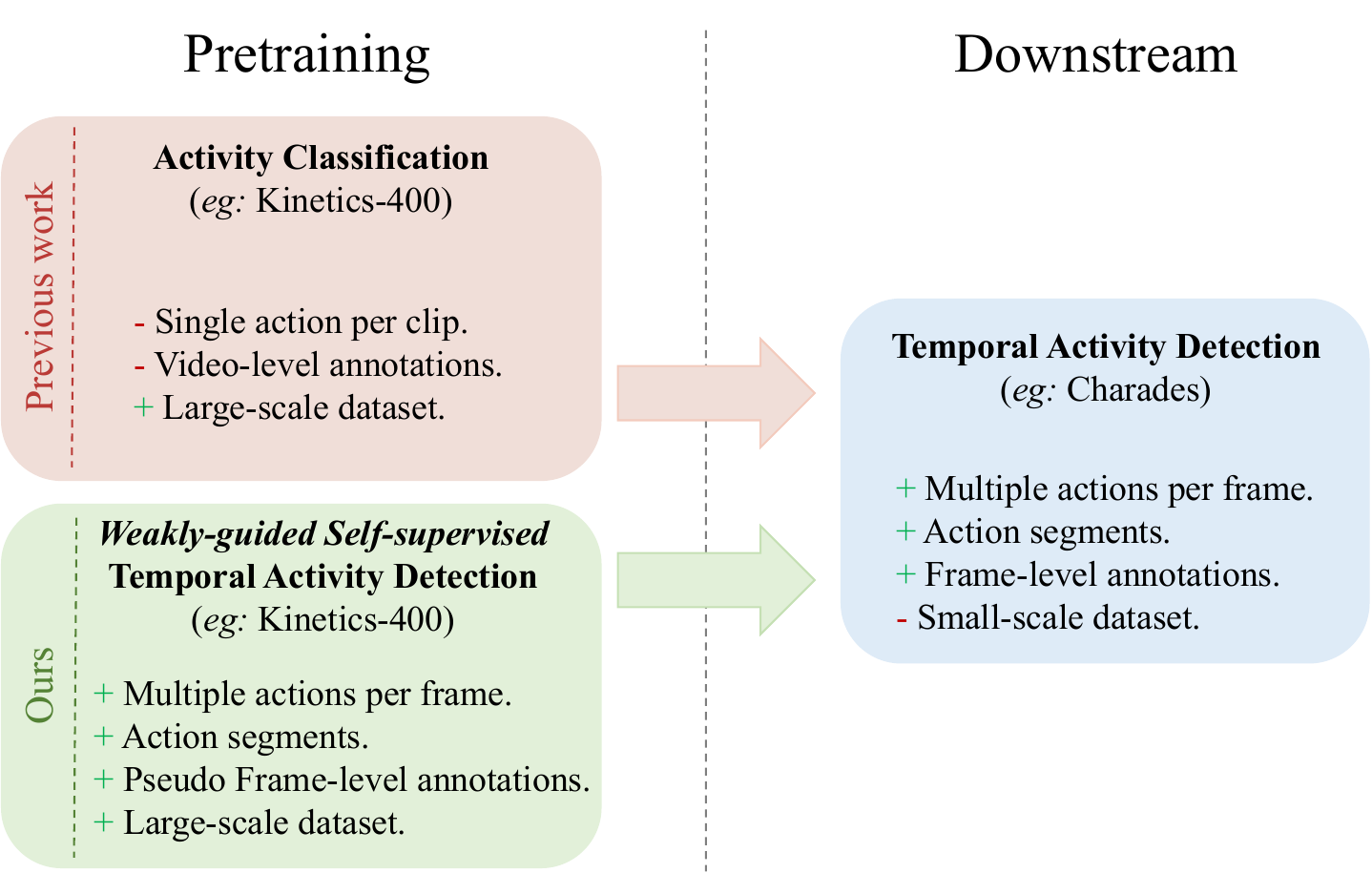}
	\caption{\textbf{Our \ch{\textit{weakly-guided self-supervised}} pretraining strategy:} Previous work on temporal activity detection are usually pretrained on large-scale activity classification datasets ($e.g.$, Kinetics-400 \cite{carreira2017quo}). However, there is a disparity between pretraining and downstream tasks, which hurts the detection performance. \ch{To bridge this gap, we propose a new self-supervised pretext task (detection) which leverages already-available weak labels (classification) to introduce frame-level pseudo labels, multi-action frames and action segments, similar to downstream. In fact, we design a detection pretraining task on large-scale classification data, without extra annotations.} %
	}
	\label{fig:intro}
\end{figure}

Even though pretraining generally helps downstream tasks, the amount of boost depends on the compatibility of the pretrained task and the downstream task \cite{abnar2021exploring}. The pretraining task (or distribution) should be as close as possible to the downstream task (or distribution) to achieve the highest possible gain. However, in a traditional pretraining pipeline, such compatibility may not always be an option. We only have a few large-scale labeled datasets limited to general tasks such as classification. Hence, models for most downstream tasks are usually pretrained in a classification task on either ImageNet-1K \cite{deng2009imagenet} (image domain) or Kinetics-400 \cite{carreira2017quo} (video domain), which often leaves a disparity between pretraining and downstream tasks.

For instance, in temporal activity detection--- which is defined as predicting (one or more) activity classes per frame--- we have the same observation: although pretraining on activity classification improves downstream detection performance, it is limited by the disparity between tasks. As a model can learn to aggregate temporal information when pretraining for activity classification (looking at the bigger picture), it may not be well-suited to do downstream activity detection, which is fine-grained and requires the model to retain temporal information as much as possible (looking at the composition of atomic actions). 
To address this issue, multiple previous work have proposed specific temporal \cite{piergiovanni2018learning, piergiovanni2019temporal, kahatapitiya2021coarse} or graphical \cite{ghosh2020stacked, mavroudi2020representation} modeling in the downstream to capture aspects not seen in the pretraining data, such as long-term motion, human-object interactions, or multiple overlapping actions in fine detail. However, it can be difficult for such finetuning techniques to alleviate the data disparity effectively.

\begin{figure}[t]
	\centering
	\includegraphics[width=0.9\linewidth]{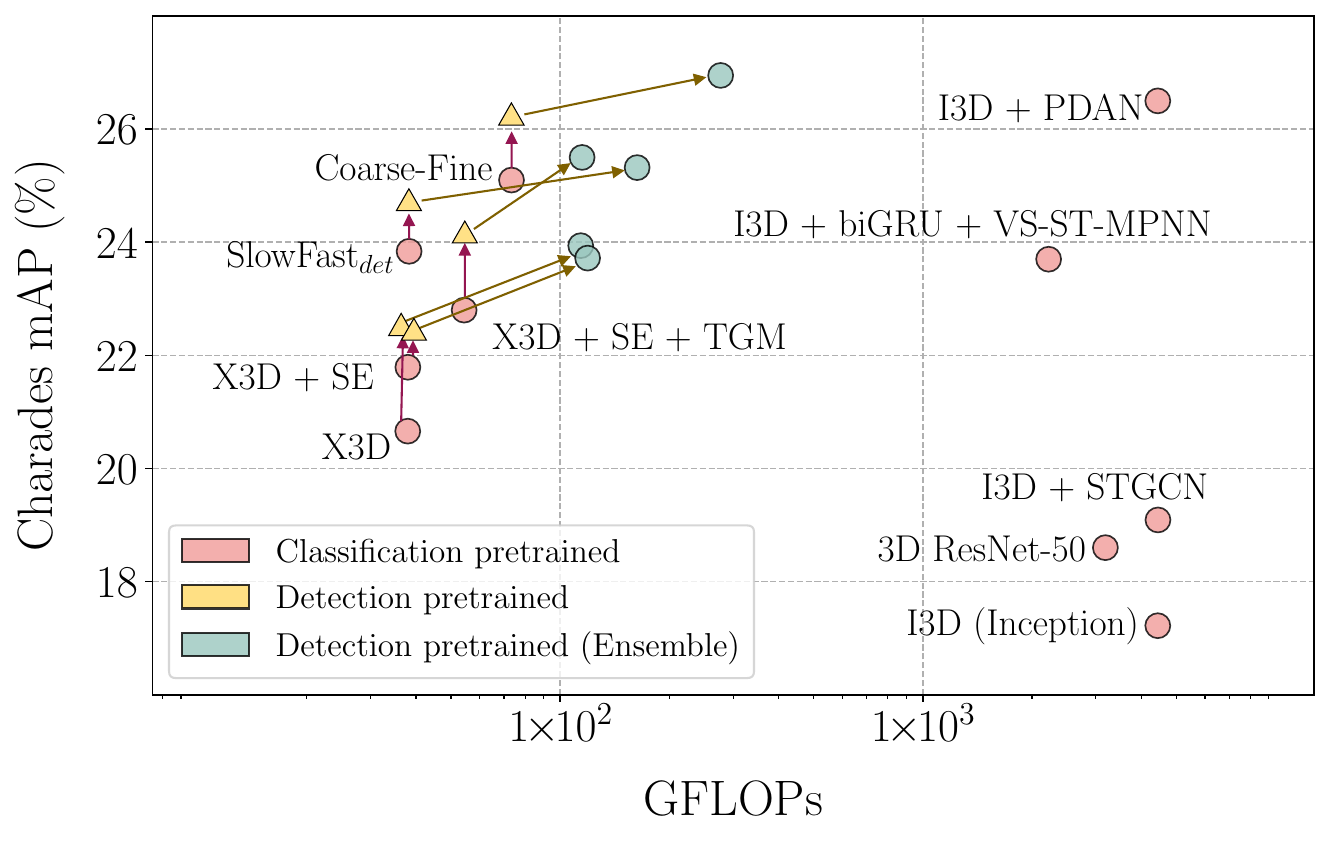}
	\caption{\textbf{Performance comparison} between models pretrained for classification and the proposed weakly-guided self-supervised detection, on downstream Charades \cite{sigurdsson2016hollywood} activity detection setting.  Representative models pretrained for detection, using \textit{Volume Freeze}, \textit{Volume MixUp} and \textit{Volume CutMix} achieve significant performance boosts over their classification pretrained counterparts. Relative improvement is shown as Classification-pretrained $\rightarrow$ Detection--pretrained $\rightarrow$ Detection-pretrained (Ensemble). Model names are shown for Classification pretrained versions in space (red circles). }
	\label{fig:improvement}
\end{figure}

In this work, we propose a \textit{weakly-guided self-supervised} pretraining method for activity detection, using large-scale classification data with \textbf{no extra annotations}. We augment pretraining data to capture fine-grained details 
and use detection as the pretraining (or pretext) task --- a step closer to bridging the gap with downstream detection (see \fref{fig:intro}). Specifically, we first extend weak video-level labels of classification clips to create pseudo frame-level labels. Then, we propose three self-supervised augmentation techniques to generate multi-action frames and action segments within a clip. Namely, we introduce \textit{Volume Freeze}, \textit{Volume MixUp} and \textit{Volume CutMix}. Volume Freeze creates a motion-less segment within a clip introducing segmented actions, whereas Volume MixUp and Volume CutMix seamlessly merge multiple clip segments into one, which tries to mimic the downstream data distribution of multiple actions per frame. Based on the augmented data, models are pretrained on an activity detection task. Our evaluations validate the benefits of the proposed pretraining strategy on multiple temporal activity detection benchmarks such as Charades~\cite{sigurdsson2016hollywood} (see \fref{fig:improvement}) and MultiTHUMOS~\cite{yeung2018every}, with multiple models such as X3D, SlowFast and Coarse-Fine. We further investigate the extent of the detection-pretrained features in our ablations and, recommend when and how to use them best.

\ch{Our method leverages weak labels 
during pretraining, having downstream settings unchanged. Also, we design a pretext task based on augmentations similar to the work in self-supervision.  %
Considering the traits of both domains, we term our work as \textit{weakly-guided self-supervision}.}

\section{Related Work}
\label{sec:related}

\paragraph{Video understanding:} Spatio-temporal (3D) convolutional architectures (CNNs) are commonly used for video modeling \cite{tran2014c3d, carreira2017quo, xu2017r}. Among these, multi-stream architectures fusing different modalities \cite{simonyan2014two, feichtenhofer2016convolutional} or different temporal resolutions \cite{feichtenhofer2019slowfast} have achieved state-of-the-art results. To improve the efficiency of video models, Neural Architecture Search (NAS) has also been explored recently in \cite{ryoo2019assemblenet, feichtenhofer2020x3d}. Multiple other directions either try to take advantage of long-term motion \cite{yue2015beyond, varol2017long, piergiovanni2018learning}, graphical modeling \cite{zhao2021video, mavroudi2020representation}, object detections \cite{baradel2018object, zhou2019grounded} or attention mechanisms \cite{nawhal2021activity, chang2021augmented, fan2021multiscale} to improve video understanding.

\paragraph{Fine-grained activity prediction:} Making predictions per frame is significantly challenging compared to activity classification (i.e., making predictions per video). It has two flavors: (1) Temporal Activity Localization (TAL) which predicts activity proposals: boundaries and corresponding classes, assuming continuity of actions \cite{shou2016temporal, escorcia2016daps, buch2017sst, yeung2016end, shou2017cdc, zhai2021action, tirupattur2021modeling, liu2021acsnet, guo2022uncertainty}, and (2) Temporal Activity Detection which explicitly predicts classes per frame \cite{piergiovanni2019temporal, kahatapitiya2021coarse, dai2021pdan}. We focus on the latter. Datasets for such tasks provide frame-level annotations with possibly multiple classes per frame \cite{caba2015activitynet, sigurdsson2016hollywood, yeung2018every}.

\paragraph{Limited Supervision:} This includes unsupervised \cite{sener2018unsupervised,kukleva2019unsupervised,gong2020learning}, self-supervised \cite{jain2020actionbytes,chen2020action}, weakly-supervised \cite{sun2015temporal} or semi-supervised \cite{ji2019learning} settings, based on the level of annotations used \cite{chen2022semi}. Self-supervision in particular, explores two directions: pretext tasks \cite{misra2016shuffle, wei2018learning, purushwalkam2020aligning, zhukov2020learning, recasens2021broaden} or contrastive learning \cite{he2020momentum, chen2020simple, chen2021exploring}.

Prior work on temporal activity localization have explored limited supervision either during pretraining \cite{zhang2022unsupervised, xu2021boundary, alwassel2021tsp, xu2021low}, or the downstream \cite{richard2017weakly, nguyen2018weakly, liu2019weakly, yu2019temporal, liu2019completeness, shi2020weakly}. We focus on pretraining, defining a pretext task (as in self-supervision) which also depends on video-level weak annotations to do fine-grained predictions (as in weak-supervision). We keep the downstream settings unchanged, with full supervision. Our formulation however, is with the flavor of frame-level predictions (activity detection), rather than predicting temporal proposals with boundaries and class labels (TAL). Thus, ours is orthogonal to above work on pretraining, but can be complementary to those on downstream finetuning.

\section{\ch{\textit{{Weakly-guided Self-supervised}}} Pretraining}
\label{method}

\begin{figure}[t]
	\centering
	\includegraphics[width=1\linewidth]{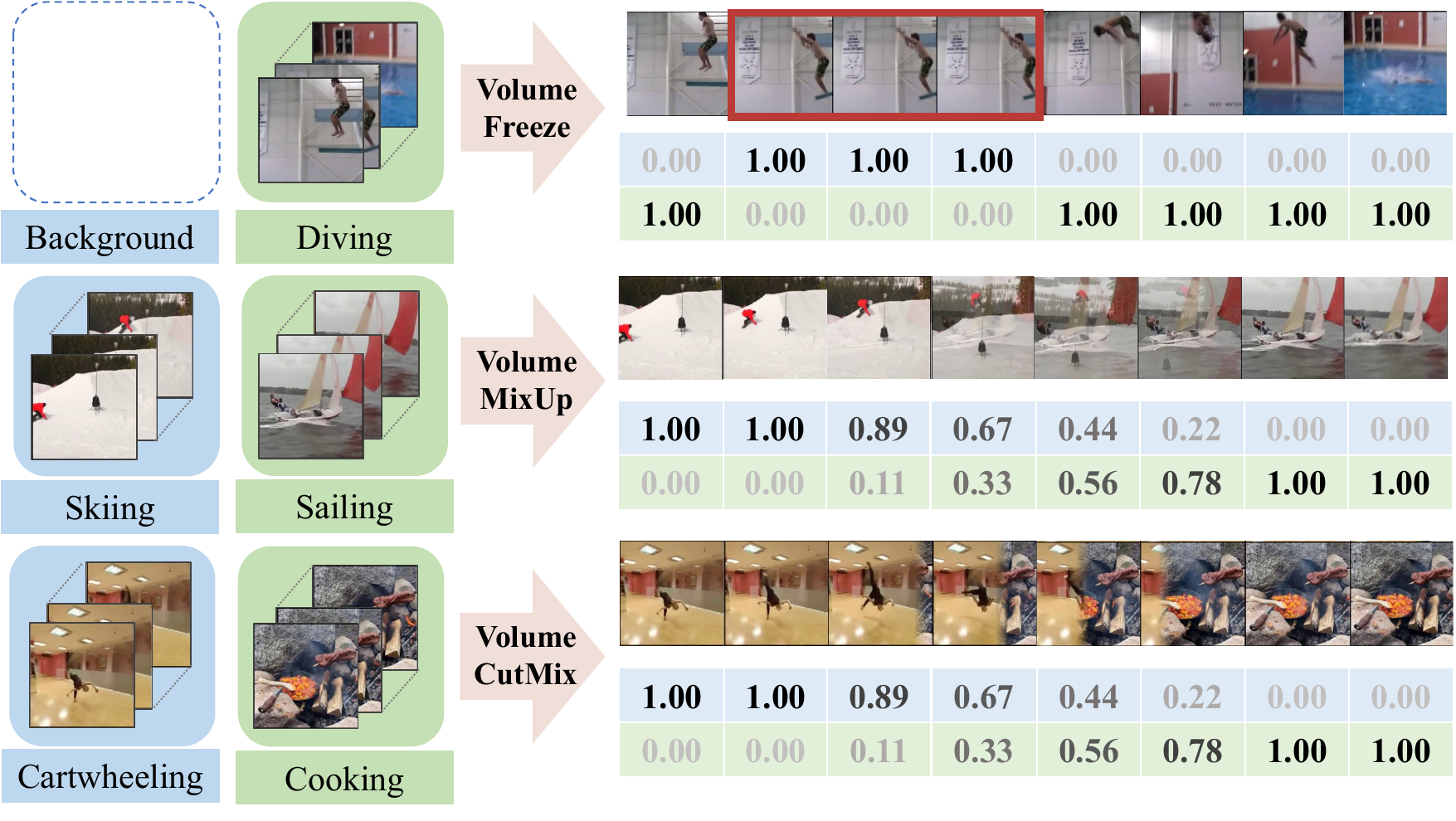}
	\caption{\textbf{Volume Augmentations for our \ch{\textit{weakly-guided self-supervised}} detection pretraining:} \textit{Volume Freeze}, \textit{Volume MixUp} and \textit{Volume CutMix}. We first extend video-level labels (of single-action videos from Kinetics-400 \cite{carreira2017quo}) into every frame, creating frame-level pseudo labels. Next, to introduce action segments and multi-action frames similar to downstream detection, we propose the above three augmentation strategies. Volume Freeze stops the motion of a video segment, creating a background segment (assuming no action can be performed without motion). Hard-labels are assigned for action and background accordingly. Volume MixUp and CutMix introduce a seamless spatio-temporal (random) transition between two clips inspired by similar ideas in image domain \cite{zhang2017mixup, yun2019cutmix}. Here, labels are weighted to create soft-labels based on the alpha values or the area of each frame, respectively. Augmented frames are best viewed zoomed-in.}
	\label{fig:aug}
\end{figure}

We introduce a self-supervised pretraining task for activity detection, which leverages already-available weak labels in large-scale classification datasets. This idea is primarily motivated based on removing the disparity between classification pretraining and downstream detection. Almost all the temporal activity detection works are pretrained for classification on large-scale datasets such as Kinetics-400 \cite{carreira2017quo}. This is because (1) video models need large-scale data to mitigate overfitting during training, and (2) detection annotations (frame-level) are too expensive to collect for a large enough dataset. Even with such classification-based pretraining at scale, the performance on downstream detection task is unsatisfactory. One reason for this is the complexity of the downstream task: predicting fine-grained activity classes per frame is challenging. Also, it can be partially attributed to the striking difference in tasks (and data distributions) during pretraining and downstream detection. As shown in \fref{fig:intro}, pretraining videos in general (eg: Kinetics-400) have only a single action per clip with video-level annotations, whereas, in a downstream detection task (eg: Charades), usually a model needs to predict multiple actions per each frame. It means that although such classification-based pretraining leveraged large-scale labeled data for training, the inherent bias which comes with it acts as a limiting factor for the downstream performance. 

\ch{We try to bridge this gap by proposing a \textit{weakly-guided self-supervised} pretraining task that closely resembles the downstream task. It shows similarities to both weak- (as we leverage weak labels) and self-supervision (as we design a pretext task based on augmentations).} Specifically, we introduce frame-level pseudo labels followed by multi-action frames and action segments through a set of data augmentation strategies. By doing so, we benefit from the scale of data, while having a similar data distribution (in terms of having overlapping and segmented actions) as downstream detection. Next, we will introduce our pseudo labeling, volume augmentations, and how we combine these ideas.

\subsection{Frame-level Pseudo Labels}

Downstream detection is about fine-grained predictions of activity classes, which requires frame-level annotations to train. However, large-scale classification datasets used for pretraining contain video-level annotations. For instance, we consider commonly-used Kinetics-400 \cite{carreira2017quo}, which contains a \textit{single} action per clip with a video-level label. As we wish to design a pretraining task that closely-resembles downstream detection, we generate frame-level labels from the available video-level labels, by replicating the same label for every frame. Such labels can be noisy because not every frame in a clip may contain the annotated single video-level action. However, we know such clips do not contain any additional actions, at least in the context of the original action categories. It is worth noting that we do not create new labels, thus no extra annotation effort is spent generating frame-level pseudo labels for classification data. 

\ch{One may also consider a pretraining dataset such as ActivityNet \cite{caba2015activitynet} with multiple actions per clip, instead of Kinetics-400 \cite{carreira2017quo} with a single action. In such a setting, an off-the-shelf action proposal generator can be used to get such pseudo frame-level labels for the proposed pretraining. However, in this paper, we consider Kinetics pretraining as commonly-used in most prior work.}

\subsection{Volume Augmentations}

Based on the frame-level pseudo labels, we design a self-supervised pretext task for detection on the pretraining data. The idea here is to introduce action segments and multi-action frames similar to the downstream data. To do this, we propose three augmentation methods specifically for video data (i.e., spatio-temporal volume): (1) Volume Freeze, (2) Volume MixUp and, (3) Volume CutMix. Next, we will explain these concepts in detail.

\subsubsection{Volume Freeze:}

\begin{figure}[t]
	\centering
	\includegraphics[width=0.9\linewidth]{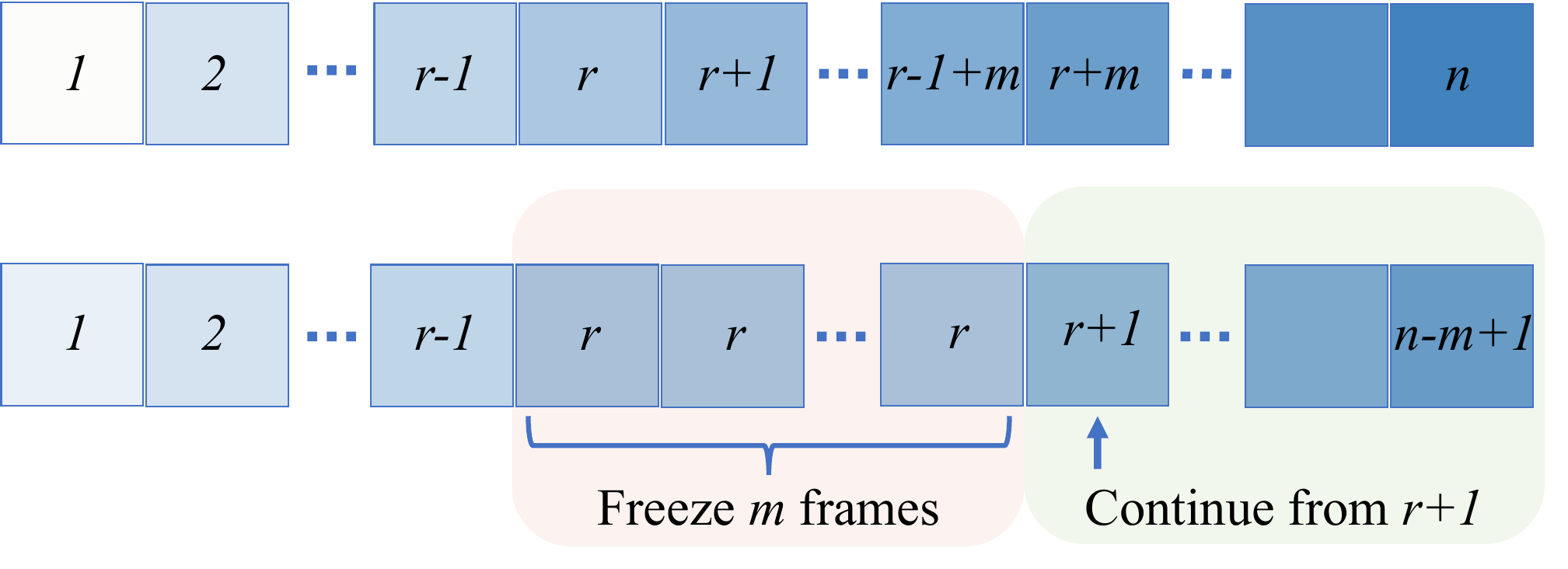}
	\caption{\textbf{Volume Freeze:} Given an input clip of length $n$, a randomly selected frame $r$ is replicated for a random $m$ duration and appended in place. Overflowing frames from the end of the clip ($t>n$) are discarded. Labels are hard labels: either action or background. Frame number is shown here with each frame.}
	\label{fig:vf}
\end{figure}

Since downstream data contains multiple action segments per clip, we want to introduce the notion of action segments in pretraining data as well. However, the videos in the pretraining dataset (Kinetics-400) contain only a single action per clip, in which, it is a challenge to have such segments. 
Our solution here is to create an motion-less (background) segment within a clip. We do this by randomly selecting a frame in a given clip, and replicating it for a random time interval (or number of frames). We call this `Background'. Such background segments are appended to the original clip at the corresponding frame location, maintaining the temporal consistency as much as possible. We label the frozen segment with a \textit{new} background label (zero-label) assuming it does not depict the original action, without any motion. Although this is a strong assumption (i.e., some actions can be classified based on appearance only, without motion), it allows the model to differentiate motion variations, giving a notion of different action segments. Volume Freeze augmentation is shown in \fref{fig:aug} (top) and elaborated \fref{fig:vf}. It can be denoted as follows,
{\small
\begin{align*}
    \text{VF}(v) &= \text{concat}(v[1:r-1], \{v[r]\}^m, v[r+1:n-m+1]), \\
    \text{VF}(l) &= \text{concat}(l[1:r-1], \{0\}^m, l[r+1:n-m+1]),
\end{align*}
}%
where $\text{VF}(v)$ and $\text{VF}(l)$ denote the augmented video and associated label in Volume Freeze. Also, $v$ and $l$ correspond to a given video clip of length $n$ and its frame-level pseudo label (one-hot), respectively. We freeze a frame for random $m$ times (denoted by $\{\cdot\}^m$) at a random temporal location $r \in [1,n-1]$, where $m \in [2,n-r+1]$, and we concatenate it to the original clip to create an augmented clip of the same original length $n$, discarding overflowing frames. \ch{This guarantees that our model does not benefit from seeing more frames compared to baseline. Also, the information loss from discarding frames is not significant, as our clip-sampling already has a significant randomness.} The labels for the augmented clip are created accordingly, where we have zero labels for the frozen segment, and original frame-level labels elsewhere. We further experiment with freezing multiple segments within a clip, which has a limited gain.

\subsubsection{Volume MixUp:}
\label{subsec:mu}

\begin{figure*}[t]
	\centering
	\includegraphics[width=0.8\linewidth]{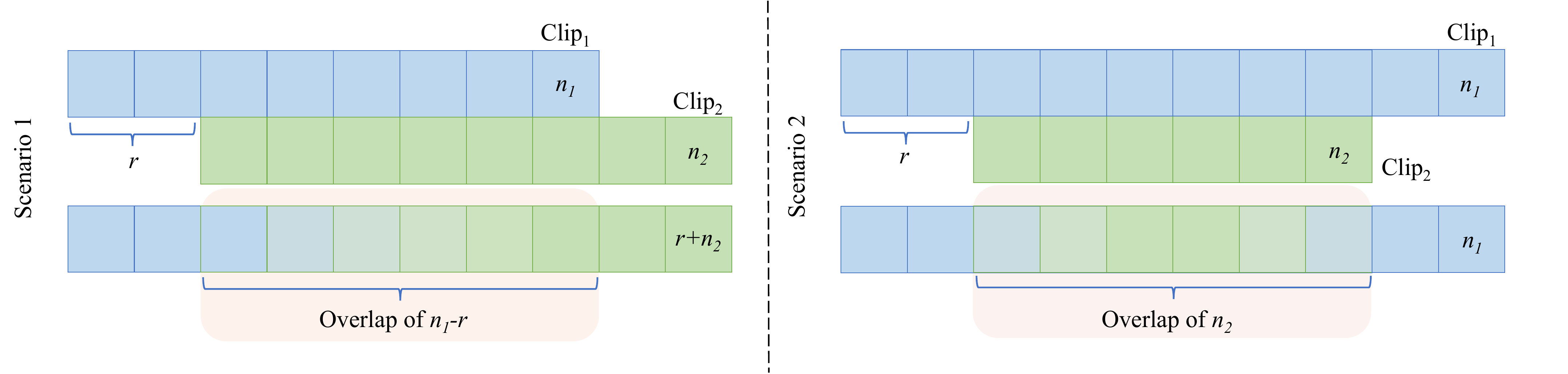}
	\caption{\textbf{Volume MixUp:} Given two input clips of length $n_1, n_2$, one clip is randomly shifted by $r$ to create a random overlap. When mixing, a seamlessly-varying alpha mask is applied in the overlapping region so that we have smooth transitions between clips. Soft-labels are created based on the alpha values. There can be two cases based on clip lengths $n_1, n_2$ and the random shift $r$: scenario 1 (top-left): Clip$_1\rightarrow\;$Clip$_2$, or, scenario 2 (top-right): Clip$_1\rightarrow\;$Clip$_2\rightarrow\;$Clip$_1$. Clip length is shown here at the end of each clip. \ch{Alpha mask is also used to weight clip labels accordingly.}}
	\label{fig:vm}
\end{figure*}

\begin{figure*}[t]
	\centering
	\includegraphics[width=0.8\linewidth]{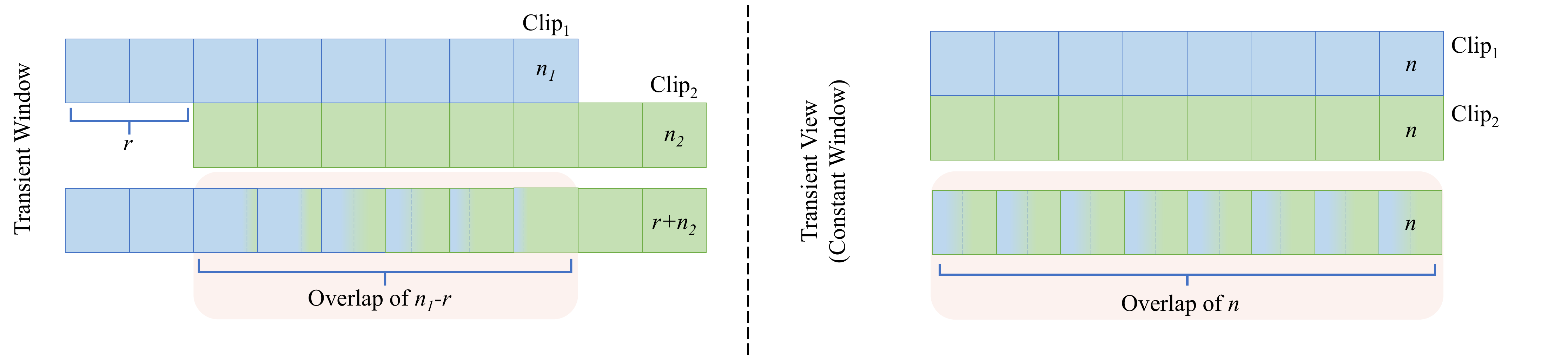}
	\caption{\textbf{Volume CutMix:} We have two settings: (1) Transient Window (top-left) and, (2) Transient View (top-right). In Transient Window, random relative shift $r$ is given similar to Volume MixUp. Smooth transition between clips is achieved when the transient window is moving from left to right (this setting can have the same two scenarios as in Volume MixUp). In Transient View, we have constant windows (half-sized) looking at transient views of the content inside (i.e., the content of each frame is moved inside the corresponding window with time, in addition to the natural motion of the clip). %
	}
	\label{fig:vc}
\end{figure*}

With Volume MixUp, we introduce multi-action frames to pretraining clips, which originally have a single action per clip. More specifically, we combine randomly selected two clips with a random temporal overlap, so that the overlapping region contains two actions per frame. This is inspired by the MixUp operation in image domain \cite{zhang2017mixup}. However, here we focus more on preserving the temporal consistency in Volume MixUp when combining two clips, by having seamlessly varying temporal alpha masks for each clip. It means, we have a smooth transition from one clip to the other within the temporal overlap. The labels for each clip are weighted with the corresponding temporal alpha mask to create soft labels. Such an augmented example with Volume MixUp is given in \fref{fig:aug} (middle) and elaborated in \fref{fig:vm}. This can also be denoted as,
{\small
\begin{align*}
    \text{VM}(v_1,v_2)[t] &= \alpha[t] \cdot v_1[t] + (1-\alpha[t]) \cdot v_2[t-r],\\
    \text{VM}(l_1,l_2)[t] &= \alpha[t] \cdot l_1[t] + (1-\alpha[t]) \cdot l_2[t-r],
\end{align*}
}%
for two video clips $v_1$ and $v_2$ of length $n_1$ and $n_2$ respectively. $v_i[t]$ and $l_i[t]$ denote the $t$-th video frame and its corresponding one-hot labels, and $\alpha[t]$ represents the scalar alpha values at time $t$ for mixing frames. Both clips are temporally padded to accommodate corresponding lengths $n_1, n_2$ and random shift $r$. \ch{The seamless temporal alpha mask for the overlapping region is defined as,}
{\small
\begin{equation*}
\label{eq:alpha}
\alpha[t]=
\begin{dcases}
\mathsf{T}_{[0,1]}(\frac{n_1-t}{n_1-r}) &\text{if}\; n_2+r\geq n_1,\\
\mathsf{T}_{[0,1]}(\frac{|n_2+2r-2t|}{n_2}) & \text{otherwise,}\\
\end{dcases}  
\end{equation*}
}%

The \textit{truncation} operator $\mathsf{T}_{[0,1]}(\cdot)$ clips the mask values within the range of $[0,1]$. It is defined in detail in appendix. %
This makes $\alpha[t]$ to be a piecewise linear function w.r.t. $t$. 
In scenario 1 ($n_2+r\geq n_1$), the augmented clip transit as Clip$_1 \rightarrow$ Clip$_2$, whereas in scenario 2, it works as Clip$_1 \rightarrow$ Clip$_2 \rightarrow$ Clip$_1$. It depends on the clip lengths $n_1, n_2$ and the random shift $r$. More details are in the Appendix. \ch{The two-clips are selected randomly (without any constraints), and hence the resulting mixed-up clip may contain artifacts. However, such randomness helps to generalize better, as also seen in \cite{zhang2017mixup}.}

\subsubsection{Volume CutMix:}
\label{subsec:cm}

Similar to Volume MixUp, we introduce multi-action frames with Volume CutMix. Here, given two clips, we define an overlapping region and assign a seamlessly changing spatial window for each clip within this region. This is inspired by CutMix \cite{yun2019cutmix} operation in image domain. In Volume CutMix however, we focus on a seamless transition between clips in time. We introduce two strategies for Volume CutMix: (1) Transient Window and (2) Transient View (Constant Window). See \fref{fig:aug} (bottom) and \fref{fig:vc}. 

\vspace{2mm}
\noindent\textit{\underline{Transient Window}}: This is closely-related to our Volume MixUp. Given two clips, we insert a random relative shift $r$ to create a random overlapping region. Clips are temporally padded at the ends to accommodate different clip lengths and shift. This can have the same two scenarios as before, depending on $n_1, n_2$ and $r$. However, rather than defining a scalar alpha mask per frame, now we define a 2D spatial window $\mathbf{M}$ as a mask, which changes seamlessly in time, within the overlapping region. The soft-labels for the overlapping region are weighted based on the area of each window. For convenience, we define the two windows based on a moving vertical plane as shown in \fref{fig:vc}. In between two windows, we have a short but smooth spatial transition, instead of a hard spatial boundary. This operation can be denoted as,
{\small
\begin{align*}
    \text{VC}(v_1,v_2)[t] &= \mathbf{M}[t] \odot v_1[t] + (1-\mathbf{M}[t]) \odot v_2[t-r],\\
    \text{VC}(l_1,l_2)[t] &= |\mathbf{M}[t]| \cdot l_1[t] + (1-|\mathbf{M}[t]|) \cdot l_2[t-r]\;,
\end{align*}
}%

where $\mathbf{M}[t]$ (defined below) is the spatial mask at time $t$. $v_i$ and $l_i$ represent a clip and the corresponding one-hot label . The symbols $\odot$ and $|\cdot|$ mean Hadamard (element-wise) product and \textit{area} of the mask (defined as the average of all its elements), respectively. More details are in the Appendix. 

\vspace{2mm}
\noindent\textit{\underline{Transient View:}} In this setting, we keep the window size constant for each clip (half of the frame) within the overlapping region (not random, but $n$ in this case). For each window to cover the spatial range of each clip, we move each clip within the constant window from left-to-right, in time. This artificial movement is introduced in addition to the natural motion in each clip. We have a constant clip length and no random shift in this case, since a zero-padding in only one-half of a frame may cause problems for convolution kernels. With the same notations as before, the augmented clip and labels can be denoted as,
{\small
\begin{align*}
    \text{VC}(v_1,v_2)[t] &= \mathbf{M} \odot v_1[t] + (1-\mathbf{M}) \odot v_2[t],\\
    \text{VC}(l_1,l_2)[t] &= 0.5 \cdot l_1[t] + 0.5 \cdot l_2[t].
\end{align*}
}%

\ch{The spatial mask $\mathbf{M}$ defines a vertical plane to split each frame within the overlapping region into two windows. The location of this vertical plane ($w_t$) can either depend on $\alpha[t]$ (in Transient Window) or be constant (in Transient View).}

\subsection{Combining Augmentations}
\label{combineAug}

In the previous subsections, we defined the components of our pretraining scheme: namely, frame-level pseudo labeling and volume augmentations. 
When combining augmentations, we use either (1) joint training or (2) model ensembling. 

\vspace{2mm}
\noindent In \textit{\underline{Joint training}}, we combine the three augmentations during training. \ch{A simpler setting is to apply only a single randomly-selected augmentation per clip (referred to as Joint train - single). Or else, we can apply up to all 3 augmentations per clip with a random probability (referred to as Joint train).} Although the latter strategy seems flexible, applying multiple of the proposed augmentations on a given sample can create confusing inputs, which are hard to train with. 

\vspace{2mm}
\noindent In \textit{\underline{Model ensembling}}, we apply only a single selected augmentation among the proposed Volume Freezing, MixUp, and CutMix during training. At inference, we combine predictions coming from such separate models trained with each augmentation. By doing so, we can combine the benefits of each augmentation, without worrying about the input confusion at training. However, this incurs more compute requirement at inference, compared to a jointly trained single model. \ch{For fair comparison, we always report the numbers for joint training (i.e., same compute budget) alongside ensembles.}

\section{Experiments}
\label{sec:results}

To validate the benefits of our proposed method, we pretrain on commonly-used Kinetics-400 \cite{carreira2017quo} and evaluate on rather-complex Charades \cite{sigurdsson2016hollywood} and MultiTHUMOS \cite{yeung2018every} for downstream detection, using the efficient video backbone X3D \cite{feichtenhofer2020x3d}. In addition to applying the proposed augmentations at the input level, we also run a few experiments with manifold augmentations \cite{verma2019manifold}, where each augmentation method is applied to the feature maps at a random depth of the network.

\subsection{Kinetics-400 Detection Pretraining}

By default, we initialize with our backbone X3D-M (medium) with checkpoints provided in original work \cite{feichtenhofer2020x3d}, \ch{as in common-practice for activity detection. This allows shorter pretraining schedules and better convergence for both our method and baseline.} We pretrain X3D for 100k iterations with a batch size of 64 and an initial learning rate of 0.05 which is reduced by a factor of 10 after 80k iterations. We use a dropout rate of 0.5. From each clip, we sample 16 frames at a stride of 5, following the usual X3D training setup. During training, first, each input is randomly sampled in [256, 320] pixels, spatially cropped to 224$\times$224, and applied a random horizontal flip. Next, we extend the labels to every frame as we described earlier, and apply one of the proposed volume augmentations to a batch of input clips. %

\ch{It is important to note that both our method and baseline are always pretrained for the \textit{exact} same number of iterations (i.e., gradient steps) and see a similar amount of data. Although, Volume MixUp and CutMix combines multiple clips per datapoint, each clip has a partial visibility, and each datapoint has the same number of total frames. This results in the same pretraining cost (see Appendix for details).}

\subsection{Charades Evaluation}
\label{subsec:charades}

We initialize X3D \cite{feichtenhofer2020x3d} with checkpoints from our detection pretraining. From each clip, we sample 16 frames at a stride of 10 and train for 100 epochs with a batch size of 16. Initially, we have a learning rate of 0.02, which is decreased by a factor of 10 at 80 epochs. For Coarse-Fine and SlowFast$_\text{det}$, we follow the same two-staged training strategy as in \cite{kahatapitiya2021coarse}. 
We train all methods on Charades with Binary Cross-Entropy (BCE) as localization and classification losses. \ch{Our models and baselines are always trained for same number of total iterations for fair comparison.} At inference, we make predictions for 25 equally-sampled frames per each input in the validation set, which is the standard Charades localization evaluation protocol \cite{sigurdsson2016hollywood} followed by all previous work. Also, it is important to note that the original evaluation script from the Charades challenge scales the Average Precision for each class with a corresponding class weight. However, in our ablations, we report the performance on predictions for every frame, which gives a more fine-grained evaluation without class-dependent weighting. Performance is measured using mean Average Precision (mAP).

\begin{table}[t!]
	\centering
	\tablestyle{1.8pt}{1.}
	\fontsize{9}{11}\selectfont
		\begin{tabular}{l|c|c|c|r}
			\multicolumn{1}{c|}{\multirow{2}{*}{Model}} & \multicolumn{1}{c|}{\multirow{2}{*}{Mod.}} & \multicolumn{2}{c|}{Pretrain}  & \multirow{2}{*}{$\;$mAP (\%)} \\
			{} & {} & cls. & det. & {} \\
			\shline
			{Two-stream I3D (Carreira et al.)} & {R+F} & \checkmark & {} & 17.22 \\
			3D ResNet-50 (\citeauthor{he2016deep}) & {R} & \checkmark & {} &   18.60 \\
			{STGCN (\citeauthor{ghosh2020stacked})} & {R+F} & \checkmark & {} & 19.09 \\
			{VS-ST-MPNN (Mavroudi et al.)} & {R+O} & \checkmark & {} & 23.70 \\
			{MS-TCT (\citeauthor{dai2022mstct})} & {R} & \checkmark & {} & 25.40 \\
			{PDAN (\citeauthor{dai2021pdan})} & {R+F} & \checkmark & {} & \underline{26.50} \\
			
			\hline
			\multirow{2}{*}{{X3D (\citeauthor{feichtenhofer2020x3d})}} & \multirow{2}{*}{R} & \checkmark & {} &   20.66 \\
			{} & {} & {} & \checkmark & (22.36) {\textbf{23.94}} \\
			\hline
			\multirow{2}{*}{{SE$^*$ (Piergiovanni et al.)}} & \multirow{2}{*}{R} & \checkmark & {} & {21.79}  \\
			{} & {} & {} & \checkmark & (22.24) {\textbf{23.92}} \\
			\hline
			\multirow{2}{*}{{TGM + SE$^*$ (Piergiovanni et al.)}} & \multirow{2}{*}{R} & \checkmark & {} & {23.84}  \\
			{} & {} & {} & \checkmark & (24.11) {\textbf{25.50}} \\
			\hline
			\multirow{2}{*}{SlowFast$_{\text{det}}^*$ (\citeauthor{feichtenhofer2019slowfast})} & \multirow{2}{*}{R} & \checkmark & {} & 22.80  \\
			{} & {} & {} & \checkmark & (24.73) {\textbf{25.32}} \\
			\hline
			\multirow{2}{*}{{Coarse-Fine (Kahatapitiya et al.)}} & \multirow{2}{*}{R} & \checkmark & {} & 25.10 \\
			{} & {} & {} & \checkmark & (26.19) {\underline{\textbf{26.95}}} \\

	\end{tabular}
	\caption{\textbf{Performance on Charades} \cite{sigurdsson2016hollywood}. We report the performance (mAP), input modalities used (R: RGB, F: optical flow or O: object), and the pretraining method: classification (cls.) or the proposed detection (det.). These results correspond to the original Charades localization evaluation setting (i.e., evaluated on evenly-sampled 25 frames from each validation clip). Model ensembles trained with our detection pretraining significantly outperform their counterparts, consistently. Coarse-Fine achieves a new state-of-the-art performance of $26.95\%$ mAP even with RGB modality only, when pretrained with our proposed method. Improved results from our pretrained ensembles are in \textbf{bold} and \ch{joint-trained single-models are within $(\cdot)$}. The best performance from each pretraining is \underline{underlined}. Model variants with X3D backbone are denoted with $^*$.}
	\label{tab:charades}
\end{table}

\begin{table*}[t!]\centering
	\captionsetup[subfloat]{captionskip=2pt}
	\captionsetup[subffloat]{justification=centering}
		
		\subfloat[{\textbf{Single stream} of X3D \cite{feichtenhofer2020x3d}: Each of the volume augmentations provide consistent improvements over the classification pretrained baseline. However, when combining augmentations, ensembles work best compared to joint-training which can create confusing inputs with multiple augmentations.}
		\label{tab:ablation:vol_aug-main}]{
		    \tablestyle{1.8pt}{1.}
	        \fontsize{8.8}{11}\selectfont
			\begin{tabular}{l|c|r}
			\multicolumn{2}{c|}{Pretraining method}  & mAP (\%) \\
			\shline
			 \multicolumn{2}{l|}{Baseline (cls.)} & {17.28} \\ \hline
			 \multirow{2}{*}{Ours (det.)} & Volume Freeze & {18.79} \\
			 \multirow{2}{*}{w/ single augmentation} & Volume MixUp & {19.18} \\
			 & Volume CutMix & {18.99} \\ \hline
			 Ours (det.) & Joint train & {19.11} \\
			 w/ multiple augmentations & Ensemble & {20.50} \\
	        \end{tabular}
	        }\hspace{2.5mm}
		\subfloat[{\textbf{Multi-stream} SlowFast$_{\text{det}}$ \cite{feichtenhofer2019slowfast}/ Coarse-Fine \cite{kahatapitiya2021coarse}: Here, we see an interesting observation. Even though detection pretrained models are consistently better as single-stream networks (eg: either Coarse/Slow or Fine/Fast), when combined as multi-stream networks, performance varies. We further investigate why this happens in Appendix. Model ensembles give consistent improvements as expected.}
		\label{tab:ablation:coarse-fine-main}]{
			\tablestyle{1.8pt}{1.}
	        \fontsize{8.8}{11}\selectfont
			\begin{tabular}{l|c|c|r}
			\multicolumn{1}{c|}{\multirow{2}{*}{Model}} & Coarse/ & Fine/ & \multirow{2}{*}{Two-stream} \\
			& Slow & Fast & \\
			\shline
			Slow$_{\text{det}}$ (cls.) - Fast$_{\text{det}}$ (cls.) & {17.49} & {17.28} & {20.31} \\
			Slow$_{\text{det}}$ (VC) - Fast$_{\text{det}}$ (VC) & {18.60} & {18.99} & ({+0.91}) {21.22} \\ \hline
			Coarse (cls.) - Fine (cls.) & {18.13} & {17.28} & {23.29} \\
			Coarse (VC) - Fine (VC) & {18.85} & {18.99} & (-0.46) {22.83} \\ 
		    Coarse (VC) - Fine (cls.) & {18.85} & {17.28} & {(+0.28) 23.57} \\ \hline
		    Coarse (Ensemble) - Fine (cls.) & {-} & {-} & {24.29} \\ 
		    Coarse (Ensemble) - Fine (cls. + Ensemble) & {-} & {-} & {\underline{24.61}} \\
	        \end{tabular}
	        }\hspace{1mm} 

		\caption{\textbf{Ablations on Charades} \cite{sigurdsson2016hollywood} with our volume augmentations in single or multi-steam models. Each augmentation gives performance boosts, and best combined as ensembles. Detection pretrained models do not show gains as good as baselines at different temporal resolutions or in temporal aggregation. This is discussed in detail in Appendix. Here, We show the performance in mean Average Precision (mAP) for fine-grained predictions (i.e., making decisions per every frame rather than evenly-sampled 25 frames from each validation clip).
		}
		\label{tab:ablations-main}
\end{table*}

\paragraph{Results:}
\label{subsubsec:main_results}
We report the performance of state-of-the-art methods comparing their pretraining strategy in \tref{tab:charades}. These numbers are for the Charades standard evaluation protocol \cite{sigurdsson2016hollywood}. We see a clear improvement from the model ensembles pretrained with the proposed detection task across multiple methods. The vanilla X3D \cite{feichtenhofer2020x3d} backbone without any additional modeling achieves the biggest relative improvement of $+3.28\%$ mAP. Detection pretraining also helps any lightweight temporal modeling on top of pre-extracted features as in super-events \cite{piergiovanni2018learning} with a $+2.13\%$ mAP and in TGM \cite{piergiovanni2019temporal} with a $+1.66\%$ mAP improvement. Finally, we see the benefits in fully end-to-end trained multi-stream networks such as SlowFast$_\text{det}$ ($+2.52\%$ mAP) and Coarse-Fine Networks \cite{kahatapitiya2021coarse} ($+1.85\%$ mAP). \ch{We also show the performance of our joint-trained single models, for fair comparison under the same compute budget. Our models consistently outperforms baselines. Note that even though our detection ensembles are compute-heavy compared to baselines, they are still an order-of-magnitude efficient compared to prior state-of-the-art PDAN \cite{dai2021pdan}.}
SlowFast$_\text{det}$ here is a variant of original SlowFast \cite{feichtenhofer2019slowfast}, with X3D \cite{feichtenhofer2020x3d} backbone, adopted for detection in \cite{kahatapitiya2021coarse}. We show the performance vs. compute trade-off graph in \fref{fig:improvement}. %

\paragraph{Ablations:} In \tref{tab:ablations-main}, we discuss the benefit of each augmentation, both separately and combined, followed by an interesting observation in multi-stream models. Each of our volume augmentation provide consistent gains, with $+1.51\%$ mAP in Volume Freeze, $+1.90\%$ mAP in Volume MixUp and $+1.71\%$ mAP in Volume CutMix. When combining augmentations, if we apply multiple of them to a given input, it may result in confusing frames. Rather, different augmentations can be complementary when used as ensembles, giving $+3.22\%$ mAP over the baseline (see \tref{tab:ablation:vol_aug-main}). In multi-stream models, we observe that our detection pretrained models do not show similar gains as baselines, (1) at different temporal resolutions or (2) in temporal aggregation (see \tref{tab:ablation:coarse-fine-main}). When selecting models based on this observation, we see consistent improvement. A detailed discussion on this and more ablations are included in the Appendix

\subsection{MultiTHUMOS Evaluation}
\label{subsec:multithumos}

\begin{table}[t!]
	\centering
	\tablestyle{1.8pt}{1.}
	\fontsize{9}{11}\selectfont
		\begin{tabular}{l|c|c|c|r}
			\multicolumn{1}{c|}{\multirow{2}{*}{Model}}  & \multicolumn{1}{c|}{\multirow{2}{*}{Mod.}} & \multicolumn{2}{c|}{Pretrain}  & \multirow{2}{*}{$\;$mAP (\%)} \\
			{} & {} & cls. & det. & {} \\
			\shline
			{Two-stream I3D (Carreira et al.)} & R+F & \checkmark & {} & 36.40 \\
			{TGM + SE (Piergiovanni et al.)} & R+F & \checkmark & {} & 46.40 \\
			{PDAN (\citeauthor{dai2021pdan})} & R+F & \checkmark & {} & \underline{47.60} \\
			
			\hline
			\multirow{2}{*}{{X3D (\citeauthor{feichtenhofer2020x3d})}} & \multirow{2}{*}{R} & \checkmark & {} &   37.17 \\
			{} & {} & {} & \checkmark & {(38.92) \textbf{40.88}} \\
			\hline
			\multirow{2}{*}{{TGM + SE$^*$ (Piergiovanni et al.)}} & \multirow{2}{*}{R} & \checkmark & {} & {39.16}  \\
			{} & {} & {} & \checkmark & {(41.55) \textbf{43.15}} \\
			\hline
			\multirow{2}{*}{{PDAN$^*$ (\citeauthor{dai2021pdan})}} & \multirow{2}{*}{R} & \checkmark & {} & {39.20}  \\
			{} & {} & {} & \checkmark & {(42.13) \textbf{\underline{44.35}}} \\

	\end{tabular}
	\caption{\textbf{Performance on MultiTHUMOS} \cite{yeung2018every}. We report the performance (mAP), input modalities used (R: RGB or F: optical flow), and the pretraining method: classification (cls.) or the proposed detection (det.). Model ensembling trained with our detection pretraining significantly outperform their counterparts consistently, and shows overall competitive results even with RGB modality only. Improved results from our pretrained ensembles are in \textbf{bold} and \ch{joint-trained single-models are within $(\cdot)$.} The best performance from each pretraining strategy is \underline{underlined}. Model varients with X3D backbone are denoted with $^*$.}
	\label{tab:thumos}
\end{table}

We follow the same training recipe as in Charades, starting with a checkpoint pretrained for our detection. At inference, we make predictions per every frame and report using mAP.
\paragraph{Results:}
\label{subsubsec:main_results_thumos}

In \tref{tab:thumos}, we show that the state-of-the-art models pretrained with the proposed detection, consistently outperform those trained with classification, both in vanilla backbones such as X3D \cite{feichtenhofer2020x3d} ($+3.71\%$ mAP), and in models which perform temporal modeling on-top of pre-extracted features as in TGM \cite{piergiovanni2019temporal} ($+3.99\%$ mAP) or PDAN \cite{dai2021pdan} ($+5.15\%$ mAP). PDAN, with our pretraining, significantly efficient X3D backbone and only RGB modality achieves competitive performance compared to multi-modal I3D \cite{carreira2017quo} counterparts.

\section{Conclusion}
\label{sec:conclusion}

\ch{This work introduced a new weakly-guided self-supervised pretraining strategy for temporal activity detection, leveraging already-available weak labels.} We defined a detection pretraining task with frame-level pseudo labels and three volume augmentation techniques, introducing multi-action frames and action segments to the single-action classification data. Our experiments confirmed the benefits of the proposed method across multiple models and challenging benchmarks. As takeaways, we further provide recommendations on when to use such pretrained models based on our observations.

\section{Appendix}
\label{sec:appendix}

\renewcommand{\thesubsection}{A.\arabic{subsection}}
\renewcommand{\thesection}{\Alph{section}}
\setcounter{table}{0}
\setcounter{figure}{0}
\renewcommand{\thetable}{A.\arabic{table}}
\renewcommand{\thefigure}{A.\arabic{figure}}

\vspace{2mm}
\subsection{Detailed ablations on Charades}
\label{subsubsec:ablations}

\begin{table*}[t!]\centering
	\captionsetup[subfloat]{captionskip=2pt}
	\captionsetup[subffloat]{justification=centering}
		
		\subfloat[{\textbf{Volume Freeze} with a single or two separate frozen segments. Multiple frozen segments does not give a considerable benefit. --- X3D \cite{feichtenhofer2020x3d}}
		\label{tab:ablation:vol_freeze}]{
		    \tablestyle{1.8pt}{1.}
	        \fontsize{8.8}{11}\selectfont
			\begin{tabular}{l|r}
			  \multicolumn{1}{l|}{Volume Freeze}  & mAP (\%) \\
			\shline
			 Baseline (cls.) & {17.28} \\ \hline
			 Single segment & {18.79} \\
			 Two segments & {\underline{18.83}} \\
			 \multicolumn{2}{c}{} \\ %
	        \end{tabular}
	        }\hspace{1.5mm}
		\subfloat[{\textbf{Volume MixUp} with seamless boundaries preserve temporal consistency and works better than having hard boundaries between clips. Gain from Manifold MixUp \cite{verma2019manifold} is minor. --- X3D \cite{feichtenhofer2020x3d}}
		\label{tab:ablation:vol_mixup}]{
			\tablestyle{1.8pt}{1.}
	        \fontsize{8.8}{11}\selectfont
			\begin{tabular}{l|r}
			 \multicolumn{1}{l|}{Volume MixUp}  & mAP (\%) \\
			\shline
			  Baseline (cls.) & {17.28} \\ \hline
			  Hard boundaries & {18.86} \\
			  Seamless & {19.18} \\ \hline
			  Seamless (manifold) & {\underline{19.24}} \\
	        \end{tabular}
	        }\hspace{1.5mm}
		\subfloat[{\textbf{Volume CutMix} with transient windows or transient views (with a constant window for each clip). Transient views show a slightly better performance. --- X3D \cite{feichtenhofer2020x3d}}
		\label{tab:ablation:vol_cutmix}]{
			\tablestyle{1.8pt}{1.}
	        \fontsize{8.8}{11}\selectfont
			\begin{tabular}{l|r}
			 \multicolumn{1}{l|}{Volume CutMix}  & mAP (\%) \\
			\shline
			  Baseline (cls.) & {17.28} \\ \hline
			  Transient window & {18.89} \\ 
			  Transient view & \multirow{2}{*}{\underline{18.99}} \\
			  (w/ Constant window) & \\ %
	        \end{tabular}
	        }\hspace{1.5mm}
		\subfloat[{\textbf{When combining augmentations}, ensembles work best compared to joint-training which can create confusing inputs with multiple augmentations. --- X3D \cite{feichtenhofer2020x3d}}
		\label{tab:ablation:combining_aug}]{
			\tablestyle{1.8pt}{1.}
	        \fontsize{8.8}{11}\selectfont
			\begin{tabular}{l|r}
			\multicolumn{1}{c|}{Method} & mAP (\%) \\
			\shline
			Baseline (cls.) & {17.28} \\ \hline
			\ch{Joint train - single} & {19.28} \\
			Joint train & {19.11} \\
			Ensemble & {\underline{20.50}} \\
	        \end{tabular}
	        }%

		\subfloat[{\textbf{Performance of multi-stream architectures} with streams pretrained with different methods. Here, we see an interesting observation: even though detection pretrained models are consistently better as single-stream networks (eg: either Coarse/Slow or Fine/Fast), when combined as multi-stream networks, performance varies. We further investigate why this happens in \tref{tab:ablation:downsampling} and \tref{tab:ablation:feat-agg}. Model ensembles however, give consistent improvements as expected.  --- SlowFast$_{\text{det}}$ \cite{feichtenhofer2019slowfast}/ Coarse-Fine \cite{kahatapitiya2021coarse}}
		\label{tab:ablation:coarse-fine}]{
			\tablestyle{1.8pt}{1.}
	        \fontsize{8.8}{11}\selectfont
			\begin{tabular}{l|c|c|r}
			\multicolumn{1}{c|}{\multirow{2}{*}{Model}} & Coarse/ & Fine/ & \multirow{2}{*}{Two-stream} \\
			& Slow & Fast & \\
			\shline
			Slow$_{\text{det}}$ (cls.) - Fast$_{\text{det}}$ (cls.) & {17.49} & {17.28} & {20.31} \\
			Slow$_{\text{det}}$ (VC) - Fast$_{\text{det}}$ (VC) & {18.60} & {18.99} & \textcolor{pos}{(\underline{+0.91})} {21.22} \\ \hline
			Coarse (cls.) - Fine (cls.) & {18.13} & {17.28} & {23.29} \\
			Coarse (VC) - Fine (VC) & {18.85} & {18.99} & \textcolor{neg}{(-0.46)} {22.83} \\ 
		    Coarse (VC) - Fine (cls.) & {18.85} & {17.28} & {\textcolor{pos}{(+0.28)} 23.57} \\ \hline
		    Coarse (Ensemble) - Fine (cls.) & {-} & {-} & {24.29} \\ 
		    Coarse (Ensemble) - Fine (cls. + Ensemble) & {-} & {-} & {\underline{24.61}} \\
	        \end{tabular}
	        }\hspace{1mm} 
		\subfloat[{\textbf{\ch{Training schedule and performance when starting from scratch or checkpoints.}} We always pretrain our detection models with the same schedule as baselines (same iterations $\rightarrow$ same computations $\times10^{15}$). Initializing models from checkpoints provided in original works, give faster and better convergence. --- X3D \cite{feichtenhofer2020x3d}}
		\label{tab:ablation:checkpoints}]{
			\tablestyle{1.8pt}{1.}
	        \fontsize{8.8}{11}\selectfont
			\begin{tabular}{l|c|c|r}
			\multicolumn{1}{c|}{\multirow{2}{*}{Model}} & \multirow{2}{*}{Pretraining} & \multirow{2}{*}{mAP (\%)} & Pretraining\\
			{} & {} & {} & FLOPs (P) \\
			\shline
		    X3D - & cls. & {16.02} & {91.10} \\
			{(300k from scratch)} & det. & {\underline{17.89}} & {91.10} \\ \hline
			X3D - & cls. & {17.28} & {30.36} \\
			{(100k from checkpoint)} & det. & {\underline{19.28}} & {30.36}\\
	        \end{tabular}
	        }%

		\subfloat[{\textbf{Statistics from the validation set}, which show the improvement (cls.$\rightarrow$det.) for single vs. multi-action frames, and action boundary vs. non-boundary regions (for a boundary considered with a dilation of 3). We see a consistently larger improvement in multi-action frames compared to single-action frames, as our pretraining introduces multi-action frames. Improvement on downstream from introducing boundaries in minimal. --- X3D \cite{feichtenhofer2020x3d}} \label{tab:ablation:stat}]{
			\tablestyle{1.8pt}{1.}
	        \fontsize{8.8}{11}\selectfont
			\begin{tabular}{l|r|r|r|r}
			\multicolumn{1}{c|}{\multirow{2}{*}{Pretraining}}  & \multicolumn{2}{c|}{Act. per frame} & \multicolumn{2}{c}{Boundary} \\
			  & \multicolumn{1}{c|}{=1} & \multicolumn{1}{c|}{$>$1} & \multicolumn{1}{c|}{False} & \multicolumn{1}{c}{True} \\
			\shline
			cls. & {8.63} & {18.72} & {17.34} & {16.18} \\ \hline
			
			det. (VF) & {\textcolor{neg}{-0.14}} & {\textcolor{pos}{\underline{+1.85}}} & {\textcolor{pos}{+1.41}} & {\textcolor{pos}{\underline{+1.49}}}\\
			
			det. (VM) & {\textcolor{pos}{+1.30}} & {\textcolor{pos}{\underline{+2.06}}} & {\textcolor{pos}{+1.87}} & {\textcolor{pos}{\underline{+1.94}}} \\
			
			det. (VC) $\;\;\;\;\;$ & {\textcolor{pos}{+1.16}} & {\textcolor{pos}{\underline{+1.86}}} & {\textcolor{pos}{+1.68}} & {\textcolor{pos}{\underline{+1.85}}}\\
			\end{tabular}
			} \hspace{2mm}        
		\subfloat[{\textbf{At lower temporal resolutions} (< pretrained resolution), detection pretrained models are not improved (Fine/Fast$\rightarrow$Coarse/Slow) as much as classification pretrained ones. Classification captures an overview of a clip which can be better generalized to different temporal resolutions. However, detection pretrained models still consistently outperform others. --- SlowFast$_{\text{det}}$ \cite{feichtenhofer2019slowfast}/ Coarse-Fine \cite{kahatapitiya2021coarse}} \label{tab:ablation:downsampling}]{
			\tablestyle{1.8pt}{1.}
	        \fontsize{8.8}{11}\selectfont
			\begin{tabular}{l|c|c|r}
			\multicolumn{1}{c|}{Pretraining}  & Fine/Fast& Coarse & \multicolumn{1}{c}{Slow} \\
			\shline
			cls. & {17.28} & {\textcolor{pos}{\underline{+0.85}}} & {\textcolor{smpos}{\underline{+0.21}}} \\ \hline
			det. (VF) & {18.79} & {\textcolor{neg}{-0.27}} & {\textcolor{neg}{-0.27}} \\
			det. (VM) & {19.18} & {\textcolor{smpos}{+0.12}} & {\textcolor{neg}{-0.01}} \\
			det. (VC) $\;\;\;\;\;$ & {18.99} & {\textcolor{neg}{-0.14}} & {\textcolor{neg}{-0.39}} \\
			\end{tabular}
			} \hspace{2mm}
		\subfloat[{\textbf{In temporal aggregation}, classification pretrained models perform better. We show this with the fusion module in Coarse-Fine \cite{kahatapitiya2021coarse}, which aggregates Fine features with Gaussians at a given standard deviation $\{T/32, T/16, T/8\}$. Here, if we increase the standard deviation, we temporally aggregate (dilate) more, where classification pretrained features show consistently higher improvement.  --- Coarse-Fine \cite{kahatapitiya2021coarse}} \label{tab:ablation:feat-agg}]{
			\tablestyle{1.8pt}{1.}
	        \fontsize{8.8}{11}\selectfont
			\begin{tabular}{l|c|c|r}
			\multicolumn{1}{c|}{Coarse-Fine$\;\;\;$}  & $T/32$ & $T/16$ & \multicolumn{1}{c}{$T/8$} \\
			\shline
			cls. & {22.80} & {\textcolor{pos}{\underline{+0.68}}} & {\textcolor{pos}{\underline{+0.49}}} \\ \hline
			det. (VC) & {22.84} & {\textcolor{smpos}{+0.14}} & {\textcolor{neg}{-0.01}} \\
			\end{tabular}
			} %

		\caption{\textbf{Derailed ablations on Charades} \cite{sigurdsson2016hollywood} \textbf{activity detection}, evaluating our design choices and showing when our detection pretrained models can be most beneficial (i.e., relative improvement from detection pretrained models are not as much as their counterparts at different temporal resolutions (\tref{tab:ablation:downsampling}) or strong temporal aggregation (\tref{tab:ablation:feat-agg})). Here, We show the performance in mean Average Precision (mAP) for fine-grained predictions (i.e., making decisions per every frame rather than evenly-sampled 25 frames from each validation clip). Relative changes of \textcolor{neg}{negative}, \textcolor{smpos}{postive-but-small} and \textcolor{pos}{postive} are shown in corresponding color, whereas the best performances that we highlight are \underline{underlined}. 
		}
		\label{tab:ablations}
\end{table*}
	
This section presents multiple ablations evaluating our design decisions and provides recommendations on when to use our detection pretrained models. Note that, in these experiments, we report the performance evaluated for every frame on Charades \cite{sigurdsson2016hollywood} (in contrast to the standard evaluation protocol of evaluating only on 25 frames per clip), which is measured using mAP (without class weights). This is similar to the original setting, but provides more robust and fine-grained performance metrics.

\paragraph{Number of frozen segments in Volume Freeze:} As shown in \tref{tab:ablation:vol_freeze}, Volume Freezing provides a relative improvement of $+1.51\%$ mAP over classification pretrained X3D \cite{feichtenhofer2020x3d}. By default, we consider a single random frozen segment in a given clip. Benefit from having multiple such frozen segments is minimal (only  $+0.04\%$ mAP).

\paragraph{Variations of Volume MixUp:} We consider Volume MixUp of two clips with hard or smooth (having seamlessly changing temporal alpha masks) boundaries. Among these, smooth boundaries preserve the temporal consistency better, giving a $+0.32\%$ mAP boost over the former, as shown in \tref{tab:ablation:vol_mixup}. Volume MixUp applied in a random feature-level as in \cite{verma2019manifold} is not much better (only  $+0.06\%$ mAP) than the same augmentation applied always at the input level.

\paragraph{Windowing strategies in Volume CutMix:} Among the windowing methods of Volume CutMix discussed earlier, transient view performs slightly better ($+0.10\%$ mAP) than transient window as shown in \tref{tab:ablation:vol_cutmix}.

\paragraph{Combining augmentations:} As illustrated in \tref{tab:ablation:combining_aug}, when combining the three volume augmentation methods as described previously, an ensemble of models pretrained separately works considerably better ($+1.22\%$ mAP) than a single model jointly-trained with all augmentations. \ch{In joint training, if we apply a single volume augmentation to each clip rather than multiple, it performs better ($+0.17\%$ mAP).} This is because multiple augmentations applied together may create confusing/cluttered inputs to the network. Ensembles can avoid such clutter by training separately, while complementing other models within the ensemble at inference.

\paragraph{Multi-stream methods and ensembling:} We see some interesting results when combining multiple streams of detection pretrained models. As shown in \tref{tab:ablation:coarse-fine}, SlowFast$_\text{det}$ \cite{feichtenhofer2019slowfast}, shows a clear improvement ($+0.91\%$ mAP) with two detection pretrained streams, whereas Coarse-Fine \cite{kahatapitiya2021coarse} does not ($-0.46\%$ mAP). However, when a classification pretrained model is used as the Fine stream, it works better ($+0.28\%$ mAP) than the original. This gives intriguing observations on detection models (1) at different temporal resolutions and, (2) in temporal aggregation, which are further explored in \tref{tab:ablation:downsampling} and \tref{tab:ablation:feat-agg}, respectively. When considering model ensembles, in Coarse-Fine, a classification pretrained Fine stream fused with detection pretrained Coarse stream ensembles gives a $+0.72\%$ mAP improvement over a single Coarse-Fine network. If we further include Fine stream ensembles, it gives an additional boost of $+0.32\%$ mAP.

\paragraph{Training schedule:} \ch{We always pretrain our detection models with the same training schedule as baseline classification models (in terms of both gradient iterations and computations) as shown in \tref{tab:ablation:checkpoints}. Our detection pretrained models consistently outperforms baselines. As often practiced in temporal activity detection, if we initialize our models with checkpoints provided in original works, it makes the models train faster (100k vs. 300k iterations) and converge to a better optimum.}

\begin{figure*}[]
	\centering
	\includegraphics[width=0.95\textwidth]{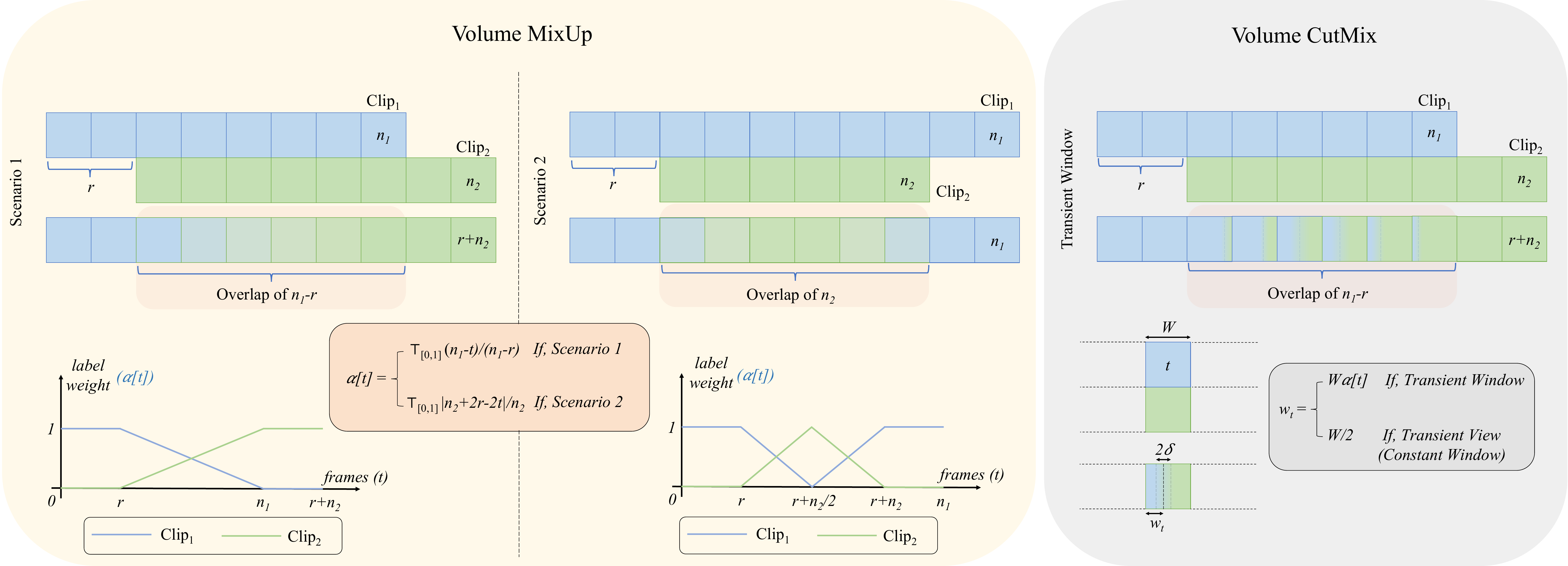}
	\caption{\textbf{Detailed view of masks} used in Volume MixUp (left) and Volume CutMix (right). In Volume MixUp, a temporal alpha mask ($\alpha[t]$) is defined, which is further visualized above (left) for both scenarios. When $n_2+r\geq n_1$ (Scenario 1), $\alpha[t]$ is defined so that the augmented clip transit from Clip$_1\rightarrow\;$Clip$_2$. Otherwise, transition happens as Clip$_1\rightarrow\;$Clip$_2\rightarrow\;$Clip$_1$. A truncation operation ($\mathsf{T}_{[0,1]}$) is applied to clip the mask value into the range of $[0,1]$. Here, the labels (one-hot) of each clip are summed with weights $\alpha[t]$ and ($1-\alpha[t]$) to create soft-labels. In Volume CutMix above (right), a spatial mask ($\mathbf{M}[t]$) is defined for each frame at time $t$, creating two windows in the overlapping region (split by a vertical plane). In Transient Window setting, the location of the vertical plane ($w_t$) depends on $\alpha[t]$ (same one as in Volume MixUp), and in Transient View (Constant Window), $w_t$ is half of the frame-width ($W$). A small spatial region of $2\delta$ is defined between widows to have a smooth spatial transition. 
The labels (one-hot) of each clip are summed with weights $|\mathbf{M}[t]|$ and $(1-|\mathbf{M}[t]|)$ to create soft-labels.  Given a matrix, $|\cdot|$ computes its ``area'' as an average of all its elements. %
	}
	\vspace{-3mm}
	\label{fig:app}
\end{figure*}

\paragraph{Statistics showing the points of improvement:} \tref{tab:ablation:stat} shows the performance boosts of each augmentation, measured under different settings to highlight what happens (1) in multi-action frames and (2) around action boundaries. Our augmentations significantly improve the mAP in a multi-action setting, as we introduce multi-action frames in pretraining. Even though we introduce action boundaries during pretraining, it does not show a contrasting change between boundary and non-boundary regions in the downstream.

\paragraph{At different temporal resolutions:} In \tref{tab:ablation:downsampling}, we consider classification and detection pretrained models at a different ($\times 4$ lower) temporal resolution, by comparing the coarse/slow stream vs. the fine/fast stream. Classification pretrained models consistently give a better relative change than the detection pretrained models (the consistent gain of Coarse/Slow w.r.t. Fine/Fast). This is because, when pretrained with classification, models can capture an overview of an input clip, allowing it to better generalize for different temporal resolutions. However, the absolute performance metrics are always better in detection counterparts.

\paragraph{In temporal aggregation:} In Coarse-Fine \cite{kahatapitiya2021coarse}, fusion module aggregates Fine features with Gaussians (at defined standard deviation). As in \tref{tab:ablation:feat-agg}, by evaluating at different standard deviations (different aggregation scales), we see that classification pretrained features give a better temporal aggregation compared to detection counterparts. This is due to the same reason mentioned above: classification features capture an overview, hence better generalize across scales. \ch{Based on these observations, we recommend using classification models in temporal aggregation or change of temporal resolution.}

\subsection{On the Truncation Operator $\mathsf{T}_{[0,1]}$}
\label{app:a}

As shown on \fref{fig:app} (left), the truncation operator $\mathsf{T}_{[0,1]}$ makes sure that the alpha mask ($\alpha[t]$) is within the range of [0,1], even in the non-overlapping region. It can be defined as,
{\small
\begin{equation*}
\label{eq:alpha}
\mathsf{T}_{[0,1]}(x)=
\begin{dcases}
1 &\text{if}\;\; x\geq 1,\\
0 &\text{if}\;\;  x< 0,\\
x & \text{otherwise},\\
\end{dcases}  
\end{equation*}
}%
where any value $x\geq1$ or $x<0$ is capped at either 1 or 0 respectively. Based on this, alpha mask $\alpha[t]$ is deifined so that the augmented clips have a smooth transition as Clip$_1\rightarrow\;$Clip$_2$ (in Scenario 1), or as Clip$_1\rightarrow\;$Clip$_2\rightarrow\;$Clip$_1$ (in Scenario 2).

\subsection{On the Spatial Mask $\mathbf{M}_t$}
\label{app:b}

Spatial mask $\mathbf{M}$ defines a vertical plane to split each frame within the overlapping region into two windows (see \fref{fig:app} (right)). The location of this vertical plane ($w_t$) can either depend on $\alpha[t]$ (in Transient Window) or be constant (in Transient View). This can be given as,
{\small
\begin{equation*}
\label{eq:alpha}
\mathbf{M}[t][:,j]=
\begin{dcases}
1 &\text{if}\;\; j < w_t-\delta,\\
0 &\text{if}\;\; j \geq w_t+\delta ,\\
\frac{w_t+\delta-j}{2\delta} & \text{otherwise},\\
\end{dcases}  
\end{equation*}
}%
and,
{\small
\begin{equation*}
\label{eq:alpha}
w_t=
\begin{dcases}
\round{W\alpha[t]} &\text{if Transient Window,}\\
\round{W/2} &\text{if Transient View,}\\
\end{dcases}  
\end{equation*}
}%
where $W$ is the width of the frame, and $\delta$ is a small value defining the smooth spatial transition between windows. $\round{\cdot}$ will round the operand to the nearest integer.

\subsection{Details on datasets}

\paragraph{Kinetics-400} \cite{carreira2017quo} is a large-scale activity classification dataset commonly-used for pretraining video models. It contains 240k training and 20k validation videos
Each clip contains a single action out of 400 human action categories, and comes with video-level annotations. Kinetics clips are usually $\app10s$ long.

\paragraph{Charades} \cite{sigurdsson2016hollywood} is a mid-scale activity classification or temporal detection dataset consisting of $\app$9.8k continuous videos with frame-level annotations of 157 common household activities. The dataset is split as $\app$7.9k training and $\app$1.8k validation videos. Each video contains an average of 6.8 activity instances, often with multiple activity classes per frame, and has longer clips averaging a duration of $\app30s$.

\paragraph{MultiTHUMOS} \cite{yeung2018every} is a small-scale dataset which contains a subset of THUMOS \cite{jiang2014thumos} untrimmed videos densely annotated for 65 different action classes. It provides action segment annotations for 413 videos, split as 200 for training and 213 for validation. On average, it contains 1.5 labels per frame and 10.5 action classes per video. When compared to Charades \cite{sigurdsson2016hollywood}, this has a significantly smaller number of videos, but each clip is longer in duration.

\bibliography{egbib}

\end{document}